# Computerization of African languages-French dictionaries

**Chantal Enguehard, Mathieu Mangeot**

Laboratoire LINA
2 rue de la Houssinière, BP 92208
F-44322 NANTES CEDEX 03 France

Laboratoire GETALP-LIG
41 rue des mathématiques BP 53
F-38041 GRENOBLE CEDEX 9 France

Email: Chantal.Enguehard@univ-nantes.fr, Mathieu.Mangeot@imag.fr

**Abstract**

This paper relates work done during the DiLAF project. It consists in converting 5 bilingual African language-French dictionaries originally in Word format into XML following the LMF model. The languages processed are Bambara, Hausa, Kanuri, Tamajaq and Songhai-zarma, still considered as under-resourced languages concerning Natural Language Processing tools. Once converted, the dictionaries are available online on the Jibiki platform for lookup and modification.

The DiLAF project is first presented. A description of each dictionary follows. Then, the conversion methodology from .doc format to XML files is presented. A specific point on the usage of Unicode follows. Then, each step of the conversion into XML and LMF is detailed. The last part presents the Jibiki lexical resources management platform used for the project.

## 1. Introduction

The work behind this paper has been done during the DiLAF project to computerize African languages-French dictionaries (Bambara, Hausa, Kanuri, Tamajaq, Zarma) in order to disseminate them widely and extend their coverage. We present a methodology for converting dictionaries from Word .doc format in a structured XML format following the Unicode character encodings and Lexical Markup Framework (LMF) standards. The natural language processing of African languages is in its infancy. It is our duty to help our colleagues from the South in this way. This requires, among other things, the publication of articles, that are a valuable resource for under-resourced languages.

Many studies have been conducted in the past in this area. However, it seemed interesting to redefine a new methodology taking into account recent developments such as the Open Document Format (ODF) or LMF standards. On the other hand, we wanted to develop the simplest possible method based solely on free and open source tools so that it can be reused by many. This method can also be used for other dictionaries and by extension, any text document (language resource at large) to be converted to XML.

## 2. Presentation of the DiLAF project

If access to computers is considered as the main indicator of the digital divide in Africa, we must recognize that the availability of resources in African languages is a handicap with incalculable consequences for the development of Information Technology and Communication Technologies (ICT). Most languages in francophone West Africa area are under-resourced (π-language) (Berment, 2004): electronic resources are scarce, poorly distributed or absent, making use of these languages difficult when it comes to introducing them into the education system and especially develop their use in writing in the administration and daily life.

Dictionaries are the cornerstone of processing natural language, be it in the mother tongue or in a foreign language. The primary function of communication is conveying meaning, yet meaning is primarily conveyed through vocabulary. As David Wilkins, a british linguist (1972) wrote “so aptly “While without grammar little can be conveyed, without vocabulary nothing can be conveyed".

Thus, to help bridge this gap, we are engaged with colleagues from North and South to improve the equipment of some African languages through, among others, the computerization of printed dictionaries of African languages.

The DiLAF project aims to convert published dictionaries into XML format for their sustainability and sharing (Streiter et al., 2006). This international project brings together partners from Burkina Faso (CNRST), France (LIG & LINA), Mali (National Resource Centre of the Non-Formal Education) and Niger (INDRAP, Department of Education, and University of Niamey).

Based on work already done by lexicographers we formed multidisciplinary teams of linguists, computer scientists and educators. Five dictionaries were converted and integrated into the Jibiki lexical resources management platform (Mangeot, 2001). These dictionaries are therefore available on the Internet[1] under a Creative Commons license:

- Bambara-French dict. Charles Bailleul, 1996 edition;
- Hausa-French dict. for basic cycle, 2008 Soutéba;
- Kanuri-French dict. for basic cycle, 2004 Soutéba;
- Tamajaq-French dict. for basic cycle, 2007 Soutéba;
- Zarma-French dict. for basic cycle, 2007 Soutéba.

1 http://dilaf.org/

The aim of these usage dictionaries is to popularize the written form of the daily use of African languages in the pure lexicographical tradition (Matoré, 1973) (Eluerd, 2000). Departing from interventionist approaches of normative dictionaries (Mortureux, 1997), the present descriptive dictionaries remain open to contributions and their online availability online will, hopefully, develop a sense of pride among users of these languages. Similarly, they will participate in the development of a literate environment conducive to increase the literacy whose low level undermines the achievements of progress in other sectors.

## 3. Presentation of the dictionaries

Four of the five dictionaries have been produced by the Soutéba project (program to support basic education) with funding from the German cooperation and support of the European Union. These dictionaries for basic education have a simple structure because they were designed for children of primary school class in a bilingual school (education is given there in a national language and in French). Most terms of lexicology, such as lexical labels, parts-of-speech, synonyms, antonyms, genres, dialectal variations, etc. are noted in the language in question in the dictionary, contributing to forge and disseminate a meta-language in the local language, a specialized terminology. The entries are listed in alphabetical order, even for Tamajaq (although it is usual for this language to sort entries based on lexical roots) because the vowels are written explicitly (this mode of classification was preferred because it is well known by children).

### 3.1. Hausa-French dictionary

The Hausa-French dictionary includes 7,823 entries sorted according to the following lexicographical order: a b ɓ c d ɗ e f fy g gw gy h i j k kw ky ƙ ƙw ƙy l m n o p r s sh t ts u w y ƴ z (République du Niger, 1999a).
They are structured with different patterns according to the part-of-speech. All entries are typographical, followed by the pronunciation (tones are marked with diacritics placed on vowels) and part-of-speech. On the semantic level, there is a definition in Hausa, a usage example (identified by the use of italics), and the equivalent in French. For a noun, the gender, feminine, plurals and sometimes dialectal variants are noted. For verbs, it is sometimes necessary to specify the degree to calculate morphological derivatives. Morpho-phonological variants of feminine and plural adjectives derivations are also written.
Example:

> **jaki** [jàakíi] *s.* **babbar dabbar gida mai kamar doki, wadda ba ta kai tsawon doki ba amma ta fi shi dogayen kunnuwa**. *Ya aza wa jaki kaya za ya tafi kasuwa. Jin.:* n. *Sg.:* **jaka**. *Jam.:* **jakai, jakuna**. *Far.:* âne

### 3.2. Kanuri-French dictionary

The Kanuri-French dictionary includes 5,994 entries sorted according to the following lexicographical order: a b c d e ə f g h i j k l m n ny o p r ɍ s sh t u w y z (République du Niger, 1999b).
The orthographic form of the entry is followed by an indication of pronunciation targeting rating tones. The part-of-speech is shown in italics, followed by a definition, a usage example, a French translation and meaning in French. Additional information may appear as variants.
Example:

> **abəɍwa** [àbə̀ɍwà] *cu.* **Kəska təngəri, kalu ngəwua dawulan tada cakkidə.** *Kəryende kannua nangaro, abəɍwa cakkiwawo.* [Fa.: ananas]

### 3.3. Soŋay Zarma-French dictionary

The Zarma-French dictionary includes 6916 entries sorted according to the following lexicographical order: a ã b c d e ẽ f g h i ĩ j k l m n ŋ ɲ o õ p r s t u ũ w y z (République du Niger, 1999d).
Each entry has an orthographic form followed by a phonetic transcription in which the tones are rated according to the conventions already set for the Kanuri. The part-of-speech specify explicitly the transitivity or intransitivity of verbs. For some entries, antonyms, synonyms and references are indicated. A gloss in French, a definition and an example end the entry.
Example:

> **ɲagas** [ɲágás] *mteeb.* ● *brusquement (détaler)* ● sanniize no kaŋ ga cabe kaŋ boro na zuray sambu nda gaabi sahã-din ● *Za zankey di hansu-kaaro no i te ɲagas*

### 3.4. Tamajaq-French dictionary

The Tamajaq-French dictionary includes 5,205 entries sorted according to the following lexicographical order: a â ă ə b c d ḍ e ê f g ğ h i î j ǰ ɣ k l ḷ m n ŋ o ô q r s ṣ š t ṭ u û w x y z ẓ (République du Niger, 1999c)
The orthographic form of the entry is followed by the part-of-speech and a gloss in French displayed in italics. For nouns, morphological information about the state of annexation is often included, the plural and gender are also explicitly stated. A definition and an example of usage follow. Other information may appear as variants, synonyms, etc. As Tamajaq is not a tonal language, phonetics does not appear.
Example:

> **əbeɣla** *sn.* mulet ♦ **Ag-anɣer əd tabagawt.** *Ibeɣlan wər tăn-tăha tămalăɣa.* anammelu.: **făkr-ejăḍ.** təmust.: **yy.** iget.: **ibəɣlan**.

### 3.5. Bambara-French dictionary

The Bambara-French dictionary of Father Charles Bailleul (1996 edition) includes more than 10,000 entries sorted according to the following lexicographical order: a b c d e ɛ f g h i j k l m n ɲ ŋ o ɔ p r s t u w y z.
This dictionary is primarily intended for French speakers wishing to improve Bambara but it is also a resource for Bambara speakers. In the words of the author himself, the dictionary "plays the role of a working tool for literacy, education and Bambara culture." To date, it can be considered as the most comprehensive dictionary of the language. It is also used by specialist of other varieties of this language like Dyula (Burkina Faso, Côte d'Ivoire) and Malinké (Guinea, Gambia, Sierra Leone, Liberia, etc.).

# 4. General conversion methodology and tools

The main objective is to convert dictionaries from a word processor format adapted for human use into XML and explicitly mark all the information so one can use them automatically in natural language processing tasks. The constraints are, on the one hand, working with free, open and multi-platform tools and on the other hand define a simple process that can be understood and then performed independently by linguists having no computer knowledge except regular expressions.

## 4.1. Conversion methodology

The conversion methodology follows these steps:

1. conversion of problematic characters to Unicode;
2. conversion of OpenOffice format to XML;
3. identification and explicit tagging of each part of information (headword, part-of-speech, etc.);
4. XML validation and manual correction of errors in the data (closed lists of values, references);
5. entries structuring following the LMF standard.

## 4.2. Tools used

The first tool allows one to edit files in original format and then convert them into XML. For this step OpenOffice (or LibreOffice) is ideal. It is free and open source. Furthermore, besides the ISO standard Open Document Format (odt), it can open many Microsoft formats (rtf, .doc as well as .docx). Finally, the XML produced is simple, especially compared to Office Open XML Microsoft.

OpenOffice has a regular expression engine for search/replace functions. This tool can be used for many steps before converting to XML. However, the search/replace may be problematic because during replacements, the boundaries of text styles can be changed (part of a word is suddenly in italics). Because we rely on styles to convert to XML, we limit the conversion to Unicode characters to keep styles intact.

Then, we need an editor to modify the files. For these operations, the XML editors are not very useful because they do not directly change the plain text with regular expressions and most are not able to edit large files like dictionaries. We recommend using a simple "raw" text editor supporting regular expressions and syntax highlighting.

For XML validation and verification steps, a web browser such as FireFox does it very nicely. It is able to detect and display the XML validation errors and can interpret CSS and XSLT style to enhance the display.

## 4.3. Incremental backups needed

The methodology intends to make backups at each stage and keep track of all search/replace operations done in order to go back when errors resulting from improper action are identified. Sometimes it happens that an error is noticed long after being made. If an error can not be corrected simply by a new search/replace, it is possible to go back from a previous version.

Despite all precautions, sometimes errors are detected very late and it is very difficult to go back. If the error can not be detected automatically, it will require manual correction. One must keep in mind that nobody is perfect and yet others even better trained had to forget the possibility to automatically correct all the errors in the conversion process.

# 5. Use of Unicode

## 5.1. Characters conversion to Unicode

Although the alphabets of languages on which we have worked (Enguehard 2009) are mainly of Latin origin, new characters needed to note specific sounds in some languages with a single character has been adopted by linguists in a series of meetings. Thus, each of the alphabets we previously presented comprises at least one of these special characters: ɓ ɗ Ǝ ɛ ɣ Ƙ ɲ ŋ ɔ ƴ. Characters composed of a Latin character and a diacritical mark were also created: âêîôûăãẽĩõũḍḷṣṭẓǧǰšɍ.

Although most of these characters are present for several years in the Unicode standard (based on the work of the ISO 10646 (Haralambous 2004)), dictionaries were written using old hacked fonts. A methodology has been defined to identify and replace the inadequate characters

**abirillu** [àbìríllù] *m.* • *avril* • annasaara handu taacanta kaŋ go marsu nda me game ra • *Abirillu, 15, 1974 no Sayni Kunce na hino sambu* • *f/b.* abirillo, abirilley

**abiyanso** [àbìyànsôo] *m.* • *aéroport* • batama kaŋ ra abiyey ga zumbu • *Tilbeeri nda Dooso sinda abiyanso kaŋ ra abiyo beeri ga zumbu* • *f/b.* abiyansa, abiyansey

**abiyo** [àbíyò] *m.* • *avion* • naarumay hari no kaŋ ra i ga boro nda jinay daŋ a ma deesi nd'ey • *Jidda no abiyey ga alfujaajey zumandi* • *him.* beene-hi • *f/b.* abiya, abiyey

**abunaadam** [àbúnăadàm] *m.* • *être humain, personne* • *f/b.* abunaadamo, abunaadamey • *di.* adamayze with the ones defined in the Unicode standard. It implies that all identified characters are recorded in a file so one can easily repeat this operation if necessary. Table 1 shows part of the list for Zarma. There is no automatic method that will detect these problematic characters. It is imperative to look at the data.

Figure 1: Excerpt of the Zarma-French dictionary in original format.

| *Origin* | *Unicode* |
|---|---|
| § | ã |
| é | ẽ |
| $ | ɲ |
| ù | ŋ |
| £ | Ɲ |

Table 1: Partial view of the Unicode correspondence table for Zarma.

### 5.2. Digraphs lexicographical order

Digraphs can be easily typed using two characters but their use changes the sort order which determines the lexicographic presentation of dictionary entries. Thus, for Hausa and Kanuri, the digraph 'sh' is located after the letter 's'. So, in the Hausa dictionary, the word "sha" (drink) is located after the word "suya" (fried), and, in Kanuri, the word "suwuttu" (undo) precedes the name "shadda" (basin).
These subtle differences can hardly be processed by software and require that digraphs appear as a proper sign in the Unicode repertoire. Some used by other languages are already there, sometimes under their different letter cases: 'DZ' (U+01F1), 'Dz' (U+01F2), 'dz' (U+01F3) are used in Slovak; 'NJ' (U+01CA), 'Nj' (U+01BC), 'nj' (U+01CC) in Croatian and for transcribing the letter " Њ " of the Serbian Cyrillic alphabet, etc.
It would be necessary to complete the Unicode standard with digraphs of Hausa and Kanuri alphabets in their various letter cases.

| | | |
|---|---|---|
| fy | Fy | FY |
| gw | Gw | GW |
| gy | Gy | GY |
| ky | Ky | KY |
| kw | Kw | KW |
| ƙy | Ƙy | ƘY |
| ƙw | Ƙw | ƘW |
| sh | Sh | SH |
| ts | Ts | TS |

Table 2: Hausa and Kanuri digraphs missing in Unicode.

### 5.3. Characters with diacritics

Some characters with diacritics are included in Unicode as a unique sign, others can only be obtained by composition.
Thus, vowels with tilde 'a', 'i', 'o' and 'u' can be found in Unicode in their lowercase and uppercase forms while the 'e' with a tilde is missing and must be composed with the character 'e' or 'E' followed by the tilde accent (U+303), which can cause renderings different from other letters with tilde when viewing or printing (tilde at a different height for example).
Letter j with caron exists in Unicode as a sign ǰ (U+1F0), but its capitalized form J̌ must be composed with the letter J and caron sign (U+30C).
The characters ẽ, Ẽ et J̌ should be added to the Unicode standard.

### 5.4. Letter case change

Word processors usually provide the letter case change function, but do not always realize it the correct way.
Thus, we found during our work that OpenOffice Writer software (3.2.1 version) fails in transforming 'ɍ' to 'R' from lowercase to uppercase or vice versa (the character remains unchanged) while Notepad++ (5.8.6 version) fails in transforming ǰ in J̌.

## 6. Conversion of the format towards XML

Figure 1 shows an excerpt of the Zarma-French dictionary in the original .odt format. All the following examples are based on this dictionary.
The Open Document Format has the great advantage of being based on XML. Instead of a conversion, we will actually retrieve the contents of the XML document, then transform it to get what we want.
A document in ODF format is actually a zip archive containing multiple files including the text content in XML. This content is stored in the "content.xml" file in the archive. To retrieve this file, some clever manipulations must be followed. On MacOs, one has to create an empty folder and then copy the .odt file inside. Then, with a terminal, the "unzip" command must be launched to unzip the file. On Windows, the .odt file extension must be changed into .zip and then the. zip archive can be opened.
The file "content.xml" can now be extracted from the archive and then renamed and placed in another location. It becomes the base file on which we will continue our work. The next step consists in editing this file with a "raw" text editor.
One may first think that since the source file "content.xml" is already in XML, it may be enough to write an XSLT stylesheet to convert the file into an XML dictionary, but the XML used in the source file is completely different from the XML targeted. Indeed, the source file comes from a word processor. It is designed for styling a document and not for structuring a dictionary entry. Therefore, it is finally easier to convert the XML file "by hands" with regular expressions than to write an XSLT stylesheet for automatically converting the source file.

## 7. Explicit tagging of the information

```
<text:span text:style-name="Phonetic_20_form">
<text:span text:style-name="T7">[àbìyànsôo]</text:
span></text:span>
```

Figure 2: Part of an entry (prononciation) in XML ODF format

This step consists in tagging explicitly all pieces of information. Each piece of information is usually distinguished from others in the original file with a different style. Figure 2 shows a part of the "abiyanso" entry (airport) in the Zarma-French dictionary. The style used to indicate the pronunciation is "Phonetic_form".

After locating the pieces of information, one must choose a set of tags to mark them.

This raises the question of the choice of the language used for tags. The choice of English as the international language of research may be privileged. But in our case, English is not a language present in our dictionaries and furthermore, it is not mastered by all linguists colleagues working on the project. The use of French solves this problem since all partners master the language. However, in the case of under-resourced languages computerization projects, we believe that it is important to encourage partners to use the words of their language to define the name of the tags. This may possibly give rise to the creation of new terms that did not exist in these languages. From a political perspective, it helps to move away from a post-colonial vision of the social status of African languages and brings new value to these languages.

The set of tag now defined, the next step is to replace the ODF markup by this new “homemade” tagset.

Simply perform search/replace operations for each type of information. For the example, the following regular expression (perl syntax) removes the tag "T7":

s/<text:span text:style-name="T7">([^<]+)
<\/text:span>//g

The second expression replaces the tag "Phonetic_form" with "ciiyaŋ":

s/<text:span text:style-name="Phonetic_20_form">
([^<+)<\/text:span>/<ciiyaŋ>$1<\/ciiyaŋ>/g

```
<sanniize>abiyanso</sanniize><ciiyaŋ>[àbìyànsôo]
</ciiyaŋ><kanandi>m.</kanandi><bareyaŋ>aéroport
</bareyaŋ><feeriji>batama kaŋ ra abiyey ga
zumbu</feeriji><silmaŋ>Tilbeeri nda Dooso sinda
abiyanso kaŋ ra abiyo beeri ga zumbu</silmaŋ>
<f>abiyansa</f><b>abiyansey</b>
```

Figure 4: Entry converted with « homemade » tags

Replacing all tags leads to the result in Figure 4.

## 8. Correction of the data

At this stage, several corrections are performed on the data.

### 8.1. XML Validation

In order to use XML tools, our file must be well formed. The manipulations of the previous step almost always introduce XML syntax errors. FireFox includes an XML parser and is also able to indicate exactly where the errors are located in the file.

Once the error is located, one has to check if it is not repeated elsewhere in the file. If this is the case, a regular expression must be written to correct the error in a systematic way instead of doing it by hand. In our case, the following regular expression can solve the problem:

s/<sanniize([^<+)<\/sanniize>/<sanniize>$1<\/sanniize>/g

The XML file is now well formed. It is then possible to manipulate it with XML tools.

**abarba** [ábàrbà] *m.* type de banane **banaana dumi no kaŋ i ga haagu ga ŋwa** *Abarba gani ŋwaayaŋ ga hin ga te boro se gunde-kuubi* budde abarbaa abarbey
**abirillu** [àbìrĭllù] *m.* avril **annasaara handu taacanta kaŋ go marsu nda me game ra** *Abirillu, 15, 1974 no Sayni Kunce na hino sambu* abirillo abirilley
**abiyanso** [àbìyànsôo] *m.* aéroport **batama kaŋ ra abiyey ga zumbu** *Tilbeeri nda Dooso sinda abiyanso kaŋ ra abiyo beeri ga zumbu* abiyansa abiyansey
**abiyo** [àbíyò] *m.* avion **naarumay hari no kaŋ ra i ga boro nda jinay daŋ a ma deesi nd'ey** *Jidda no abiyey ga alfujaajey zumandi* beene-hi abiya abiyey
**abunaadam** [àbúnăadàm] *m.* être humain, personne abunaadamo abunaadamey adamayze

Figure 3: Compact view in a browser

### 8.2. Verification of closed lists of values

The stage of verification of information taking their value in a closed list is important. Some errors come from bad handling in the previous steps, while others were present in the original file before conversion. For example, a dictionary uses parts-of-speech, a termbase uses a list of domains, etc. Make a copy of the file and keep only the values to check is a systematic approach for verification.

In the example of Figure 4, the part-of-speech marked by "kanandi" can be extracted with the following expression:

s/^*<kanandi>([^<]+)<\/kanandi>*$/$1/

The resulting list must then be sorted alphabetically. TextWrangler and Notepad + + plugin with its TextFX have the necessary commands. If the editor does not offer this option, OpenOffice Calc spreadsheet can be used. This approach is then used to quickly detect irregularities. If a value appears only once, it is very likely that this is a mistake. In the dictionary used in the examples, we corrected "alteeb" to "alteeb.", "Dah." to "dab.", "m/tsif." to "m / tsif.", etc.

### 8.3. Simple corrections

A CSS style sheet can be set to view the data directly in a browser. A compact display with a different style for each type of information helps to detect structuring errors in an entry. In the example of Figure 3, we see immediately that definition (in bold) and example (in italics) are lacking for the entry "abunaadam".

With an XSL stylesheet, one can modify the data before display like adding a unique identifier for each entry, then,

for each reference define a hypertext link to the corresponding entry. When the linguist browses the file, s/he can click on the hyperlinks to verify that the references are also entries of the dictionary like the entry "abunaadam" in Figure 3 with a reference to the entry "adamayse".

It is essential to scrutinize the data to detect some errors, even if they can be fixed automatically thereafter with regular expressions. The data visualization step is also very important from a pedagogical point of view. It allows to show the benefits of XML encoding the data, in particular that several forms (style) can be associated with the same information (data). By learning the basics of CSS, the lexicographers can modify the style sheets themselves.

## 9. Structure of the entries

The entries can now be restructured. In files coming from word processors, the data structure is usually implied. We will have to add new structural elements to move towards a more standardized structure, allowing subsequent reuse. Concerning standards, LMF (Romary et al., 2004) became an ISO standard in November 2008 (Francopoulo et al., 2009). It suits ideally our goals. As it is a meta-model and not a format, we can apply the principle of the LMF model to our entry structure and keep our tags without using the LMF syntax. The core meta-model LMF is shown in Figure 5. The object "Lexical Entry" contains a "Form" and one or more "Sense" objects.

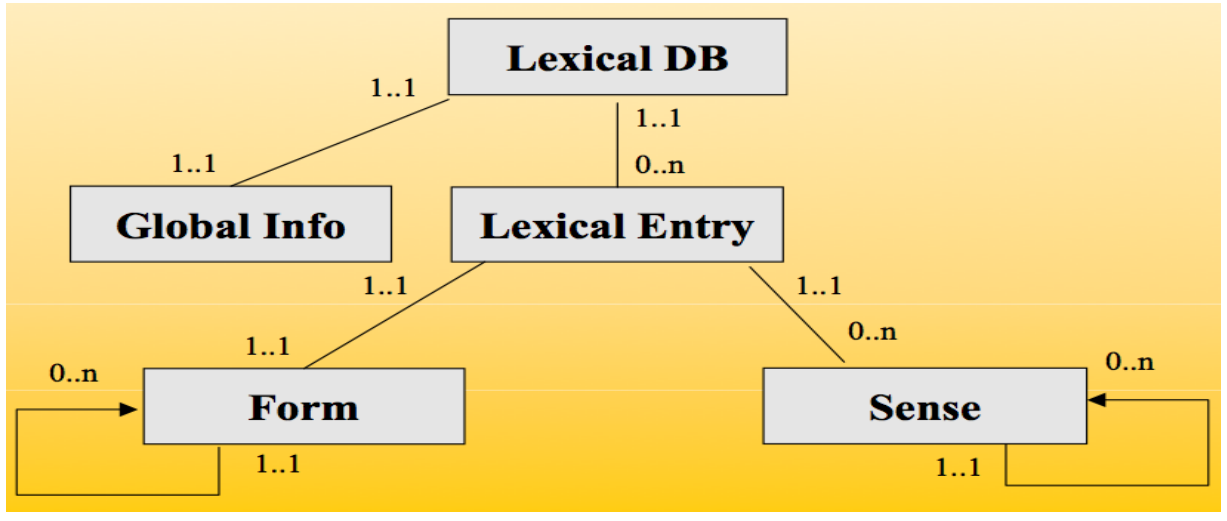

Figure 5: LMF kernel meta-model

Our lexical entries must now follow this meta-model. Figures 6 and 7 show an example of an entry before and after the addition of structuring tags. The "article" tag corresponds to the "Lexical Entry" object; the “bloc-vedette” tag correspond to the "Form" object and the "bloc-semantique" tag is the "Sense" object.

```
<sanniize>abiyanso</sanniize>
<ciiyaŋ>[àbìyànsôo]</ciiyaŋ>
<kanandi>m.</kanandi>
<bareyaŋ>aéroport<bareyaŋ>
<feeriji>batama kaŋ ra abiyey ga zumbu.</feeriji>
<silmaŋ>Tilbeeri nda Dooso sinda abiyanso kaŋ ra abiyo beeri
ga zumbu</silmaŋ>
<f>abiyansa</f><b>abiyansey</b>
```

Figure 6: Zarma entry before structuring

```
<article>
 <bloc-vedette>
  <sanniize>abiyanso</sanniize>
  <ciiyaŋ>[àbìyànsôo]</ciiyaŋ>
 </bloc-vedette>
 <bloc-grammatical>
  <kanandi>m.</kanandi>
  <f>abiyansa</f><b>abiyansey</b>
  <bloc-sémantique>
   <bareyaŋ>aéroport<bareyaŋ>
   <feeriji>batama kaŋ ra abiyey ga zumbu.</feeriji>
   <silmaŋ>Tilbeeri nda Dooso sinda abiyanso kaŋ ra abiyo beeri
ga zumbu</silmaŋ>
  </bloc-sémantique>
 </bloc-grammatical>
</article>
```

Figure 7: Entry after structuring following LMF model

```
<LexicalEntry id="abiyanso">
 <Lemma>
  <feat att="writtenForm" val="abiyanso"/>
  <feat att="phoneticForm" val="àbìyànsôo"/>
 </Lemma>
 <feat att="partOfSpeech" val="m."/>
 <Sense id="1">
  <Equivalent>
   <feat att="language" val="fra"/>
   <feat att="writtenForm" val="aéroport"/>
  </Equivalent>
  <Definition>
     <feat att="writtenForm" val="batama kaŋ ra abiyey ga
zumbu"/>
  </Definition>
  <Context>
   <TextRepresentation>
    <feat att="language" val="dje"/>
      <feat att="writtenForm" val="Tilbeeri nda Dooso sinda
abiyanso kaŋ ra abiyo beeri ga zumbu."/>
   </TextRepresentation>
  </Context>
 </Sense>
</LexicalEntry>
```

Figure 8: Zarma entry in LMF syntax

A simple XSLT stylesheet is then provided for download with each dictionary.

```
<xsl:template match="article">
   <LexicalEntry id="{sanniize}{sanniize/@lamba}">
     <xsl:apply-templates />
   </LexicalEntry>
</xsl:template>
<xsl:template match="bloc-vedette">
   <Lemma>
     <xsl:apply-templates />
   </Lemma>
</xsl:template>
<xsl:template match="sanniize">
       <feat att="writtenForm" val="{.}"/>
</xsl:template>
<xsl:template match="ciiya">
   <feat att="phoneticForm" val="{.}"/>
</xsl:template>
```

Figure 9: excerpt of the Zarma XSL stylesheet for producing LMF syntax

It converts each dictionary into the LMF syntax (see Figure 8). For more detailed information about this part, refer to (Enguehard & Mangeot, 2013).
The next step planned is to convert the resources into the Lemon format[2] and integrate them into dbnary[3] (Sérasset, 2014), the team database for linked data.

## 10. Web access via the Jibiki platform

### 10.1. Presentation of the platform

Jibiki (Mangeot et al., 2003; Mangeot et al., 2006; Mangeot, 2006) is a generic platform for handling online lexical resources with users and groups management. It was originally developed for the Papillon Project. The platform is programmed entirely in Java based on a the "Enhydra" environment. All data is stored in XML format in a Postgres database. This website mainly offers two services: a unified interface for simultaneous access to many heterogeneous resources (monolingual or bilingual dictionaries, multilingual databases, etc.) and a specific editing interface for contributing directly to the dictionaries available on the platform.
Several lexical resources construction projects used or still use this platform successfully. This is the case for the GDEF project (Chalvin et al., 2006) building an Estonian-French bilingual dictionary[4], the LexALP project about multilingual terminology on the Alpine Convention or more recently MotÀMot project on southeast Asias' languages[5]. The source code for this platform is freely available for download from the forge of the LIG laboratory[6].
An instance of the platform has been adapted specifically to DiLAF project[1] because, in addition to dictionaries, specific project information must be accessible to visitors:
- presentation of the project and partners;
- general methodology form converting published dictionaries to LMF format;
- stylesheets for different tools or tasks to be performed: tutorial on regular expressions, methodology of converting a document that uses fonts not conform to the Unicode standard to a document conforming to the Unicode standard, list of software used (exclusively open-source), methodology to monitor the project;
- presentation of each dictionary: original authors, principles that governed the construction of the dictionary, language, alphabet, structure of the lexical entries, etc.
- dictionaries in LMF format.

It is also envisaged to localize the platform for each language of the project.

2 http://lemon-model.net/
3 http://dbnary.forge.imag.fr/
4 http://estfra.ee
5 http://jibiki.univ-savoie.fr/motamot/
6 http://jibiki.ligforge.imag.fr

### 10.2. Lookup interfaces

Three different interfaces are available to the user:
- the generic lookup allows the user to lookup a word or a prefix of a word in all the dictionaries available on the platform. The language of the word must be specified.
- the volume lookup allows the user to lookup a word or prefix on a specific volume. On the left part of the result window, the volume headwords are displayed, sorted in alphabetical order. An infinite scroll allows the user to browse the entire volume. On the right part of the window, the entries previously selected on the left part are displayed.
- the advanced lookup is available for complex multi-criteria queries. For example, it is possible to lookup an entry with a specific part-of-speech, and created by a specific author. On the left part of the result window, the headwords of the matching entries are displayed, sorted in alphabetical order. An infinite scroll allows the user to browse all the matching entries. On the right part, the entries previously selected on the left part are displayed.

### 10.3. Editing process

The editor (Mangeot et al., 2004) is based on an HTML interface model instantiated with the lexical entry to be published. The model is generated automatically from an XML schema describing the entry structure. It can then be modified to improve the rendering on the screen. Therefore, it is possible to edit any type of dictionary entry provided that it is encoded in XML.
The editing process can be adapted for specific needs through levels and status. A quality level (eg: from 1 to 5 stars, an entry with 1 star is a draft and one with 5 stars is certified by a linguist) can be assigned to each contribution. Similarly, a competence level can be assigned to each contributor (1 star is a beginner and 5 stars is a certified linguist). Then, when a 3 stars level user edits a 2 stars entry, the entry level raises to 3 stars.
Status can also be assigned to entries and roles to users. For example, in order to produce a high quality dictionary, an entry must follow 3 steps: creation by a registered user, revision by a reviewer and validation by a validator.

### 10.4. Remote access via a REST API

Once dictionaries are uploaded into the Jibiki server, they can be accessed via a REST API. Lookup commands are available for querying indexed information: headword, pronunciation, part-of-speech, domain, example, idiom, translation, etc. The API can also be used for editing entries. The user must be previously registered in the website.

## 11. Conclusion

We presented a methodology for dictionaries conversion from word processing files to XML format. The DiLAF project does not stop in so good way. Before distributing dictionaries, there are still manual correction steps and possibly data addition. For example, examples of the Zarma-French dictionary will be translated into French. Once dictionaries are converted, we can then extend their coverage through a system of contribution / editing / validation that can be done online live on the Jibiki platform. The low Internet access in Africa will require us to develop alternative methods. We can then use the data as raw material to increase the computerization of these languages: morphological analysers, spell-checkers, machine translation systems, etc.

## 12. Acknowledgements

The DiLAF project is funded by the Fonds Francophone des Inforoutes of the International Organization of the Francophonie. We also thank all the linguists of the team without whom this project would not have been possible: Sumana Kane, Issouf Modi, Michel, Radji, Rakia, Mamadou Lamine Sanogo.